\pdfoutput=1

\documentclass[11pt]{article}

\usepackage[]{ACL2023}

\usepackage{times}
\usepackage{latexsym}

\usepackage[T1]{fontenc}

\usepackage[utf8]{inputenc}
\usepackage[hang,flushmargin]{footmisc}  
\usepackage{booktabs}      

\usepackage{microtype}

\usepackage{inconsolata}

\usepackage{graphicx}
\usepackage{multirow}
\usepackage{pifont}
\usepackage{xcolor}
\usepackage{subcaption}
\usepackage{float}
\usepackage{array}
\usepackage{colortbl}

\newcommand{\LN}{\linebreak\noindent}   

\newcommand*\colourcheck[1]{%
  \expandafter\newcommand\csname #1check\endcsname{\textcolor{#1}{\ding{52}}}%
}
\colourcheck{black}
\colourcheck{red}

\NewDocumentCommand{\rot}{O{90} O{0.5em} m}{\makebox[#2][l]{\rotatebox{#1}{#3}}} 

\title{Exploring the Impact of Human Evaluator Group \\ on Chat-Oriented Dialogue Evaluation}

\author{Sarah E. Finch \hspace{2em} James D. Finch \hspace{2em} Jinho D. Choi \\
  Department of Computer Science \\
  Emory University \\
  Atlanta, GA, USA \\
  \texttt{\{sfillwo, jdfinch, jinho.choi\}@emory.edu}
  }

\begin{document}
\maketitle

\begin{abstract}
Human evaluation has been widely accepted as the standard for evaluating chat-oriented dialogue systems. However, there is a significant variation in previous work regarding who gets recruited as evaluators. Evaluator groups such as domain experts, university students, and professional annotators have been used to assess and compare dialogue systems, although it is unclear to what extent the choice of an evaluator group can affect results.
This paper analyzes the evaluator group impact on dialogue system evaluation by testing 4 state-of-the-art dialogue systems using 4 distinct evaluator groups. Our analysis reveals a robustness towards evaluator groups for Likert evaluations that is not seen for Pairwise, with only minor differences observed when changing evaluator groups. Furthermore, two notable limitations to this robustness are observed, which reveal discrepancies between evaluators with different levels of chatbot expertise and indicate that evaluator objectivity is beneficial for certain dialogue metrics.
\end{abstract}
\section{Introduction}

It is common for chat-oriented dialogue modeling\LN to rely on human evaluation for comparing the performance of different dialogue systems, as automated metrics have been shown to be insufficient \cite{liu:16, deriu:22}. There is not a standard pool of human evaluators used across all works, though, as works tend to recruit their evaluators from many sources, with the most common being students \cite{zhou_learning:21, gu:22}, crowd workers \cite{liu:21, kim:21}, annotation companies \cite{song:21, cao:22}, and experts \cite{varshney:22}. 

The impact of varying evaluator characteristics has been thoroughly studied for other applications, with many indicating that evaluator characteristics have a substantial impact on the provided judgments \cite{dundes:01, vella:17}, although not for every application \cite{young:09}.
The current degree to which dialogue evaluation metrics are consistent across people of varying backgrounds is unclear due to limited work exploring this effect, which is concerning from the perspective of comparing results across works using different evaluator groups. 

This paper investigates the impact of different evaluator groups on multi-turn human evaluation of chat-oriented dialogues. 
Two popular evaluation methods are used: Likert ratings and Pairwise selections. 
For our experiments, 4 groups of evaluators are invited to provide evaluations on the same dialogue dataset: chatbot developers, professional annotators, university students with interactive point-of-view (POV), and university students with external POV. Their evaluations are then compared via dialogue-level and bot-level measures.

Our work illustrates that certain evaluation methods are impacted by the choice of evaluator group, although the degree of impact varies. In particular, we contribute 3 main findings 
that can guide future dialogue evaluations:

\begin{enumerate}
    \item Likert ratings are more stable than Pairwise selections across evaluator groups
    \item Chatbot developers produce dissimilar evaluations to less-experienced groups
    \item Objective dialogue metrics achieve better consistency among external evaluators
\end{enumerate}

\noindent Our data and analysis scripts can be accessed through our open-source project: {\small\url{https://github.com/sfillwo/DialogueEval-AnnotatorImpact}}.

\section{Related Work}
\label{sec:related_work}


There have been a handful of works that compare the evaluation agreement between different evaluator groups. \citet{ram:18} observed a high ($r = 0.93$) correlation between the 15 bot rankings of average Likert quality from Alexa Prize users and the average Likert engagingness and coherence from Amazon employees. 
\citet{venkatesh:18} claimed that they observed low agreement between their internal Amazon employee raters and Alexa Prize users for Alexa Prize chatbot conversations without providing their numerical results. 
\citet{kulikov:19} examined the differences in the average ratings provided by individual annotators for a greedy-search dialogue model and found that these averages can be dramatically different.

A few works have focused specifically on the difference between expert evaluators and non-experts. \citet{finch_towards:20} demonstrate that experts and non-experts do not agree on Likert quality ratings. Similarly, \citet{higashinaka:21} find that experts are more consistent on utterance error labeling than non-experts, although they do not compare the agreement between experts and non-experts.

\section{Data}
\label{sec:data}

\noindent For this work, we use ABC-Eval from \cite{finch:23}, a dialogue dataset containing 400 30-turn human-bot dialogues from 4 chatbots: Blender2 \cite{weston:21}, Emora \cite{finch_emora:20}, Blender-Decode \cite{nie:21}, and BART-FiD-RAG \cite{shuster:21}. 
We make use of the following common evaluation methods that are included in ABC-Eval:
\begin{itemize}
    \item \textbf{Likert rating}: human annotators provide a rating from 1 to 5 for how well the chatbot's responses fits a metric definition.
    \item \textbf{Pairwise selection}: human annotators are shown 2 dialogues and select the dialogue for which the chatbot's responses better fit the metric definition.
\end{itemize}
The 8 metrics that are included are shown in Table \ref{tab:dims}. ABC-Eval provides such evaluations from two groups: $Stu_i$ and $Sur_x$, covering two common evaluation groups of interactive students and crowdworkers, respectively, for dialogue evaluation. To cover additional common groups, we also collect evaluations from chatbot development experts and external students, denoted $Dev_x$ and $Stu_x$ respectively. Details on each evaluator group are presented next.

\begin{table}[ht]
\small
\centering\resizebox{\columnwidth}{!}{
  \begin{tabular}{|m{1.5em}|m{6cm}|}
  \hline
  \tt \textbf{Gra} & Responses are free of grammatical and semantic errors \\ \hline
  \tt \textbf{Rel} & Responses are on-topic with the immediate dialogue history \\ \hline
  \tt \textbf{Inf} & Responses produce unique and non-generic information that is specific to the dialogue context \\ \hline
  \tt \textbf{Emo} & Responses indicate an understanding of the user's current emotional state and provide an appropriate emotional reaction based on the current dialogue context \\ \hline
  \tt \textbf{Eng}  & Responses are engaging to user and fulfill the conversational goals implied by the user \\ \hline
  \tt \textbf{Con} & Responses do not produce information that contradicts other information in the dialogue \\ \hline
  \tt \textbf{Pro} & Responses actively and appropriately move the conversation along different topics \\ \hline
  \tt \textbf{Qua} & The overall quality of and satisfaction with the dialogue \\
  \hline
  \end{tabular}}
  \caption{The 8 dialogue evaluation metrics and their definitions; table adapted from \citet{finch_towards:20}.}
  \label{tab:dims}
\end{table}

\paragraph*{Interactive Students ($\mathbf{Stu_i}$)} 
Undergraduate students are recruited through email advertisements. 
The hired students receive links to the online evaluation platform to complete their assigned sessions. Each session is composed of conversing with a pair of chatbots and providing evaluations. Likert evaluations are collected on the 8 metrics after each conversation and Pairwise evaluations are collected on the 8 metrics at the end of the session. 


\paragraph*{SurgeHQ Crowdworkers ($\mathbf{Sur_x}$)}
The annotation company SurgeHQ\footnote{\href{https://www.surgehq.ai/}{\url{https://www.surgehq.ai}}} provides a pool of historically high-performing crowdworkers (hereafter, referred to as Surgers) for the ABC-Eval project. The evaluation tasks are posted as jobs on the SurgeHQ crowdworking platform and the annotation interface is identical to that used by $Stu_i$. A single Likert job consists of providing Likert ratings on 8 metrics for one dialogue. A single Pairwise job consists of providing pairwise selections between two dialogues on the 8 metrics. 


\paragraph*{Chatbot Developers ($\mathbf{Dev_x}$)} 
A group of chatbot developers involved in the development of a university chatbot are recruited from an ongoing project. None of the chatbot developers were involved in the development of any of the 4 chatbots being evaluated in this work. Their tasks follow the same format as $Sur_x$ although an internal evaluation platform is used rather than that provided by SurgeHQ. 

\paragraph{External Students ($\mathbf{Stu_x}$)}
A group of undergraduates are recruited from the same university as $Stu_i$ via email advertisements. Their task configuration is the same as $Dev_x$.

\vspace{1em}
\noindent All groups are composed of native English speakers and are shown the same instructions and interface for annotation. Evaluators in $Sur_x$, $Dev_x$, and $Stu_x$ evaluate static conversations \textit{in which they did not participate}, unlike $Stu_i$. A subset of the dialogues are doubly annotated per group, meaning that 2 human annotators provided evaluations for those dialogues. 

Table \ref{tab:data_stats} provides statistics on the evaluations from each group. It should be noted that the number of evaluators in $Stu_x$ and $Dev_x$ is much smaller than that of $Stu_i$ and $Sur_x$. Our multiple attempts to recruit participants for $Stu_x$ were met with little interest from the student population, whereas the recruitment for $Stu_i$ in \cite{finch:23} did not seem to suffer from such disinterest based on their success in obtaining such a large number of willing participants. One likely explanation for this is that the task of \textit{conversing} with dialogue models is much more compelling to human participants compared to just \textit{evaluating} human-bot dialogues. Similarly, the specialization criteria severely reduces the population from which the evaluators can be drawn from for $Dev_x$. Due to their smaller sizes, it was challenging to collect full evaluations on the dialogue dataset even over several months for both $Stu_x$ and $Dev_x$.

\begin{table}[ht]
    \small
    \centering
    \begin{tabular}{l|cccc}
    \toprule
       \bf Group & $Stu_i$ & $Stu_x$ & $Sur_x$ & $Dev_x$\\
    \midrule
       \bf Evaluators & 46 & 8 & 32 & 3 \\
    \midrule
       \bf Likert & 400 (0) & 228 (37) & 400 (108) & 177 (25) \\
       \bf Payment & $\dagger$ & \$0.50 & \$0.60 & \$0.50 \\
    \midrule
       \bf Pairwise & 200 (0) & 193 (19) & 192 (54) & 72 (11) \\
       \bf Payment & \$1.67 & \$1.00 & \$1.43 & \$1.00 \\
    \bottomrule
    \end{tabular}
    \caption{Statistics on the number of evaluators, number of evaluated dialogues (\# of doubly annotated conversations in parentheses), and compensation amount for each group. $\dagger$: Due to the interactive setup, $Stu_i$ received compensation covering Likert and Pairwise work together and only produced singly annotated dialogues.}
    \label{tab:data_stats}
\vspace{-2ex}
\end{table}




\section{Dialogue Score Agreement}

One aspect of evaluation robustness is whether the same dialogue is given the same score by different evaluators. For this, between-group interannotator agreement (IAA) acts as a measure of the impact of changing evaluator group. Higher agreement between groups signals that their dialogue-level evaluation decisions are more similar.

Following \citet{finch:23}, we use Krippendorff's alpha ($\alpha$) to measure IAA.
The between-group IAA is measured by pairing decisions made for the same dialogues across each group. The within-group IAA is calculated from the doubly annotated dialogues per group.
Table \ref{tab:agreement} shows the results. 

\begin{table}[htbp]
\centering
\resizebox{\columnwidth}{!}{%
\begin{tabular}{l|llllllll}
\toprule
 & \texttt{\textbf{Con}} & \texttt{\textbf{Emo}} & \texttt{\textbf{Eng}} & \texttt{\textbf{Gra}} & \texttt{\textbf{Inf}} & \texttt{\textbf{Pro}} & \texttt{\textbf{Qua}} & \texttt{\textbf{Rel}} \\
 \midrule
\bf Dev$_x$ / Stu$_i$ & 0.27 & 0.15 & 0.30 & \bf 0.38 & 0.05 & 0.20 & 0.34 & 0.34 \\
\bf Dev$_x$ / Stu$_x$ & \bf 0.51 & \bf 0.40 & \bf 0.42 & 0.27 & -0.30 & 0.23 & \bf 0.51 & \bf 0.45 \\
\bf Dev$_x$ / Sur$_x$ & 0.24 & 0.33 & 0.19 & 0.20 & \bf 0.16 & 0.12 & 0.33 & 0.14 \\
\bf Stu$_i$ / Stu$_x$ & 0.30 & 0.12 & 0.19 & 0.08 & -0.00 & \bf 0.32 & 0.28 & 0.16 \\
\bf Stu$_i$ / Sur$_x$ & 0.20 & 0.24 & 0.24 & 0.05 & 0.12 & 0.21 & 0.23 & 0.28 \\
\bf Stu$_x$ / Sur$_x$ & 0.30 & 0.17 & 0.16 & 0.19 & 0.06 & 0.25 & 0.27 & 0.10 \\
\rowcolor{lightgray}
\bf Dev$_x$ & 0.48 & 0.59 & 0.60 & 0.16 & 0.41 & 0.69 & 0.61 & 0.44 \\
\rowcolor{lightgray}
\bf Stu$_x$ & 0.45 & 0.17 & 0.12 & 0.42 & 0.26 & 0.13 & 0.46 & 0.05 \\
\rowcolor{lightgray}
\bf Sur$_x$ & 0.36 & 0.26 & 0.26 & 0.13 & 0.41 & 0.24 & 0.29 & 0.30 \\ 
\midrule
\bf Dev$_x$ / Stu$_i$ & \bf 0.38 & 0.54 & 0.37 & 0.28 & 0.16 & 0.21 & 0.35 & 0.36 \\
\bf Dev$_x$ / Stu$_x$ & 0.19 & \bf 0.57 & \bf 0.69 & \bf 0.32 & \bf 0.40 & \bf 0.66 & \bf 0.40 & \bf 0.40 \\
\bf Dev$_x$ / Sur$_x$ & 0.16 & 0.11 & 0.23 & 0.13 & 0.25 & 0.27 & 0.22 & 0.14 \\
\bf Stu$_i$ / Stu$_x$ & 0.13 & 0.20 & 0.24 & 0.22 & 0.09 & 0.40 & 0.32 & 0.27 \\
\bf Stu$_i$ / Sur$_x$ & 0.01 & 0.13 & 0.10 & 0.12 & 0.06 & 0.13 & 0.16 & 0.11 \\
\bf Stu$_x$ / Sur$_x$ & 0.28 & 0.21 & 0.25 & -0.04 & 0.26 & 0.25 & 0.38 & 0.17 \\
\rowcolor{lightgray}
\bf Dev$_x$ & 0.82 & 0.64 & 1.00 & 0.90 & 0.32 & 0.64 & 1.00 & 0.64 \\
\rowcolor{lightgray}
\bf Stu$_x$ & 0.27 & 0.01 & 0.07 & 0.38 & 0.07 & 0.48 & -0.15 & 0.12 \\
\rowcolor{lightgray}
\bf Sur$_x$ & 0.51 & 0.20 & 0.10 & 0.22 & 0.19 & 0.25 & 0.47 & 0.44 \\
\bottomrule
\end{tabular}}
\caption{$\alpha$ for Likert (top) and Pairwise (bottom). Gray row color indicates within-group $\alpha$'s. \textbf{Bold} indicates highest between-group $\alpha$.}
\label{tab:agreement}
\vspace{-3ex}
\end{table}




\paragraph{Overall Low Agreement} Across the metrics, the between-group agreements are rather low, rarely surpassing $\alpha=0.5$, showing that the dialogue-level judgements are rarely matched between groups. In addition, there is not an obvious difference between the agreements observed for Likert and Pairwise evaluations; thus neither evaluation procedure seems to be more robust to changing the evaluator group on the dialogue-level.

\paragraph{High Agreement Developers} For both Likert and Pairwise, the highest agreement between any two groups is achieved when one of the groups is $Dev_x$. The developer group seems to have the most in common with the other groups in terms of the specific dialogue-level judgements.



\addtocounter{footnote}{1}

\begin{figure*}
\centering
\begin{subfigure}{\textwidth}
    \includegraphics[width=\textwidth]{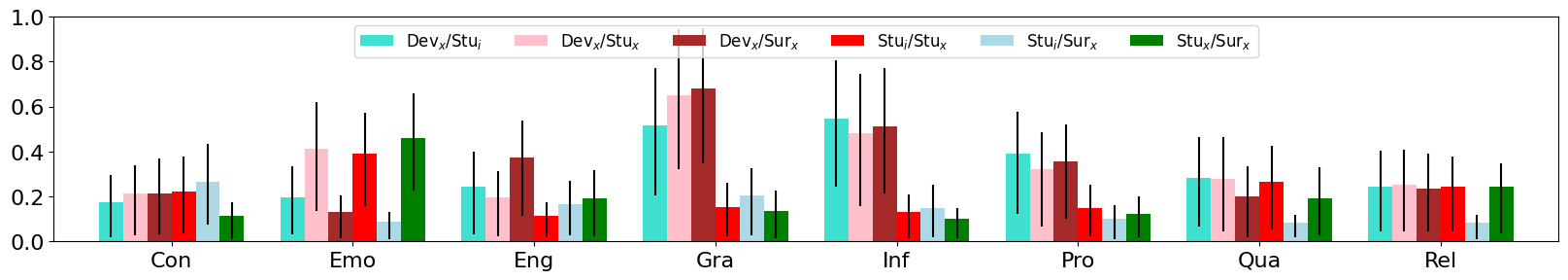}
    \caption{Likert}
    \label{fig:likert_ef}
\end{subfigure}
\begin{subfigure}{\textwidth}
    \includegraphics[width=\textwidth]{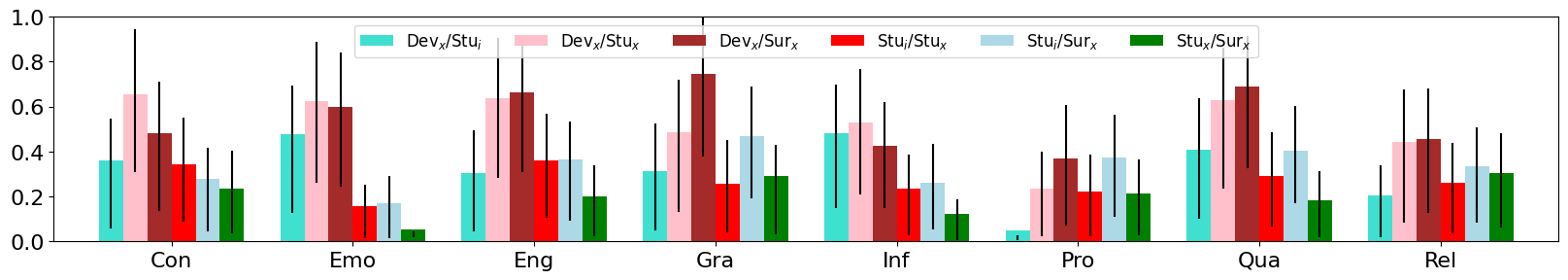}
    \caption{Pairwise}
    \label{fig:comparative_ef}
\end{subfigure}
\caption{Bot-pair effect size differences between groups (\textbf{smaller} is better). Error bars: 95\% confidence intervals.\footnotemark}
\label{fig:ef}
\vspace{-2ex}
\end{figure*}

\addtocounter{footnote}{-1}
\addtocounter{footnote}{-1}

\section{Bot Performance Analysis}
\label{sec:botanalysis}

Although dialogue-level judgements are one outcome of these evaluations, the ultimate goal of dialogue evaluation is to compare various bots to one another. Indeed, it has been seen that low dialogue-level agreement does not necessarily result in low agreement on relative bot performances when assessing evaluation strategies \cite{lee:20, ji:22}. As a result, looking at the dialogue-level judgements alone is not enough to understand the effect of changing evaluator groups and it is crucial to directly assess the relative bot performances between evaluator groups.


We first considered testing for differences in the aggregate scores for each bot produced by different evaluator groups. However, such results are not useful towards the goal of comparing the relative bot performances produced by different evaluator groups as they consider the evaluation of a bot in isolation. For example, suppose evaluator groups A and B score bot X at 3.0 and 4.0 respectively, and they score bot Y at 3.5 and 4.5 respectively. Given equal variance in their scoring distributions and sufficient sample size, testing with respect to each bot in isolation would detect that the scores produced by A and B are different when rating a particular bot, even though their evaluation of the 2 bots’ relative performance was highly analogous. 

To appropriately address the desired relative bot-pair evaluation outcomes, we instead look to utilize the effect size produced by an evaluation. Comparing the effect sizes produced by different evaluator groups directly measures evaluation results for bot-pairs, where a smaller difference in effect size signals greater similarity in the evaluations. We calculate the average difference in bot-pair effect sizes between two groups as follows:
$$
\frac{1}{N}\sum_{(b,b') \in B} |E(b_{Sur}, b'_{Sur}) - E(b_{Stu}, b'_{Stu})|
$$
$B$ is the set of bot-pairs, $N$ is the size of $B$, $b_{group}$ is a list of annotations of bot $b$ from $group$, and $E$ is the effect size function, Cohen's $d$ for Likert ratings and Cohen's $h$ for Pairwise bot-vs-all win proportions. $|\cdot|$ denotes absolute value. The results for effect size differences are shown in Figure \ref{fig:ef}.\footnote{Raw evaluation outcomes are in Appendix \ref{app:outcomes}.}

\addtocounter{footnote}{1}
\footnotetext{Bias-corrected and accelerated bootstrap confidence intervals with $k = 10,000$ Monte Carlo case resamples.} 

\paragraph{Developer Dissimilarity} 
$Dev_x$ often showcases the greatest difference relative to other evaluator groups, surpassing an effect-size difference of 0.4 for many metrics. This effect is more consistently observed for the Pairwise evaluations, although it is observed for several Likert metrics as well. 

Even though $Dev_x$ achieved some of the best agreement with other groups on the \textit{dialogue}-level, $Dev_x$ frequently produced low similarity on the \textit{bot}-level effect-sizes compared with those groups. Although this may seem contradictory, it is possible for both high dialogue-level agreement and low bot-level effect-size similarity to occur. One possible explanation is that the ratings provided by $Dev_x$ in the cases of disagreement were more extreme than those produced by the other groups, resulting in larger observed effect-sizes for $Dev_x$ evaluations and, thus, larger effect-size differences between $Dev_x$ and other groups. In other words, if $Dev_x$ produced more extreme ratings on the disagreed upon instances than the other groups, then $Dev_x$ would identify larger differences between the bots overall. This explanation is corroborated by the bot-pair effect sizes observed for each group shown in Fig. \ref{fig:ef_raw} (Appendix \ref{app:effectsize}), which shows that $Dev_x$ consistently identifies higher bot-pair effect sizes than any other group. Based on the consistency of this effect, it is likely that an evaluator's level of expertise has more of an impact on bot evaluation than other evaluator characteristics. 

\paragraph{Likert Cross-group Robustness} 
Overall, there is a lower difference in effect sizes between groups for the Likert evaluations compared to Pairwise evaluations.
This result indicates that Likert evaluations are more robust to group changes. Furthermore, the observed effect-size differences for Likert metrics are often small, rarely surpassing 0.2, between the evaluator groups (excluding $Dev_x$). As a result, the evaluation outcomes observed between these different evaluator groups are likely to manifest as only minor changes in practice.  




\paragraph{Objectivity Favors External Point-of-View} 
The smallest effect-size difference for $Con$ is seen between $Stu_x$/$Sur_x$, for both Likert and Pairwise evaluations. This is also observed for $Inf$, although the impact is more profound for Pairwise than for Likert. It is likely that these metrics with their more objective foci benefit from having external evaluators who are divorced from the conversation itself, thus reducing their subjective and emotional attachment.

\section{Limitations and Future Work}

This study was limited to 4 chatbots and 4 evaluator groups, two of which did not provide complete evaluations for the 400 total dialogues (Sec. \ref{sec:data}). Although the findings presented in this work indicate certain effects of the changing evaluator group, further work on additional group characteristics and dialogue models will aid in gaining a more complete picture of the impact of different evaluator groups on dialogue evaluation results.

\section{Conclusion}

The analyses presented in this work provide insight into the effects that switching evaluator groups can have on the results of dialogue model evaluations. The results support a recommendation for utilizing Likert ratings due to their higher stability across evaluator groups. Furthermore, if the ultimate goal is to understand how laypeople would evaluate a particular bot, we would discourage the use of chatbot developers for evaluation based on the observed dissimilarities and encourage the use of external evaluators for those metrics that are more objective. 

\bibliography{custom}
\bibliographystyle{acl_natbib}

\newpage
\clearpage
\appendix



\section{Evaluation Outcomes}
\label{app:outcomes}

\subsection{Bot-pair Effect Size}
\label{app:effectsize}

\begin{figure}[htb]
\centering
\begin{subfigure}{\columnwidth}
    \includegraphics[width=\textwidth]{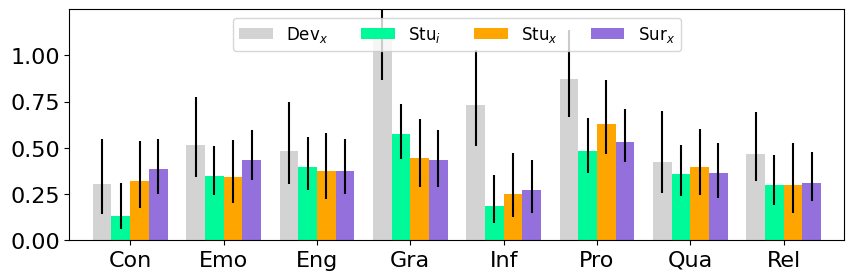}
    \caption{Likert}
    \label{fig:ef_raw_likert}
\end{subfigure}
\begin{subfigure}{\columnwidth}
    \includegraphics[width=\textwidth]{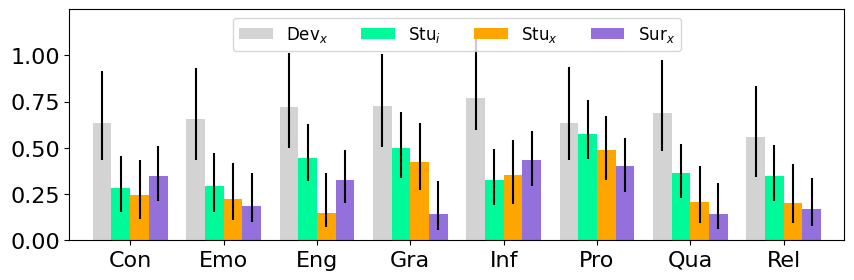}
    \caption{Pairwise}
    \label{fig:ef_raw_comparative}
\end{subfigure}
\caption{Bot-pair effect sizes for each group. Error bars denote bias-corrected and accelerated 95\% confidence intervals with
k =10,000 Monte Carlo case resamples.}
\label{fig:ef_raw}
\end{figure}

Figure \ref{fig:ef_raw} shows the average of the absolute value bot-pair effect sizes for all of the evaluator groups from the collected evaluations. 


\subsection{Likert Outcomes}
\label{sec:likert_outcomes}

\begin{figure}[htb]
\centering
\begin{subfigure}{0.75\columnwidth}
    \includegraphics[width=\columnwidth]{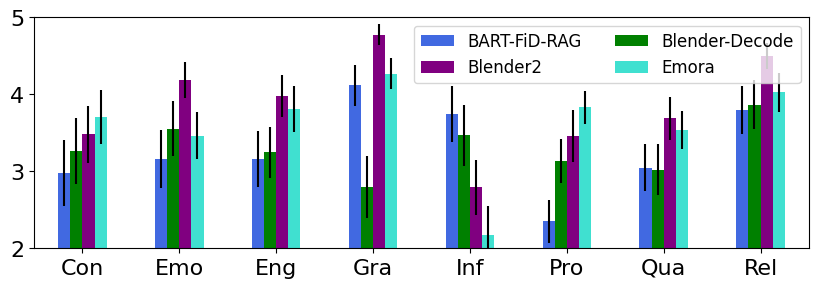}
    \caption{$Dev_x$ evaluations}
    \label{fig:developer_likert_dialogue}
\end{subfigure}
\begin{subfigure}{0.75\columnwidth}
    \includegraphics[width=\columnwidth]{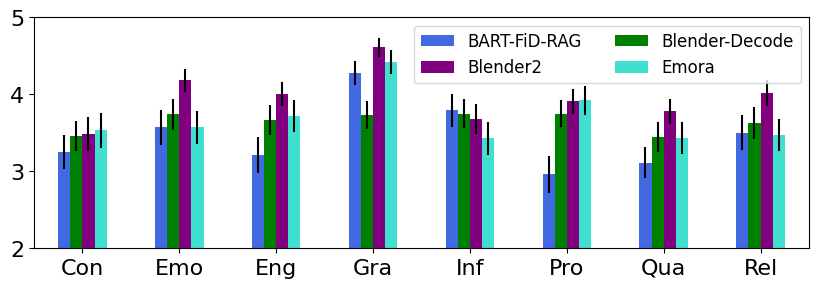}
    \caption{$Stu_i$ evaluations}
    \label{fig:interactive_likert_dialogue}
\end{subfigure}
\begin{subfigure}{0.75\columnwidth}
    \includegraphics[width=\columnwidth]{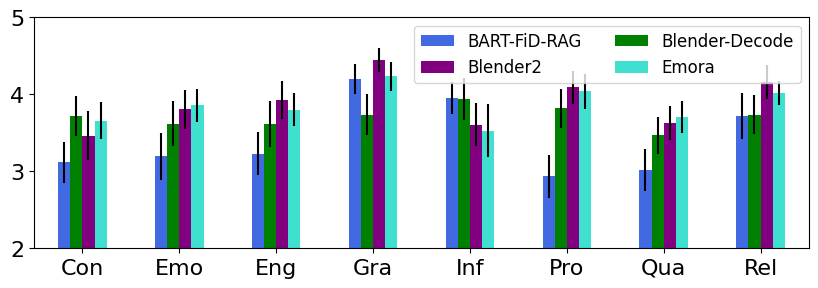}
    \caption{$Stu_x$ evaluations}
    \label{fig:nondeveloper_likert_dialogue}
\end{subfigure}
\begin{subfigure}{0.75\columnwidth}
    \includegraphics[width=\columnwidth]{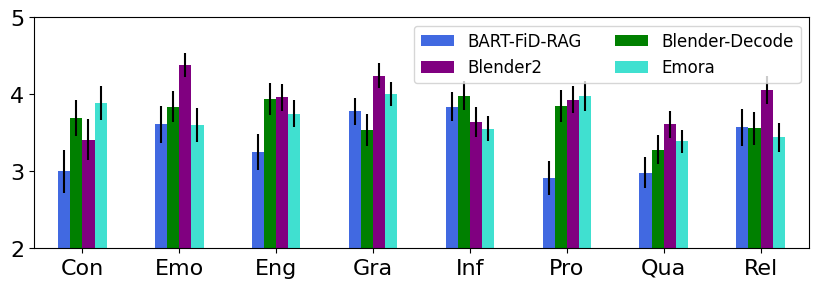}
    \caption{$Sur_x$ evaluations}
    \label{fig:external_likert_dialogue}
\end{subfigure}
\caption{Average Likert ratings for the 4 bots, with 95\% Student's t confidence intervals.}
\label{fig:likert}
\end{figure}


To provide a holistic overview of the Likert evaluation results, Figure \ref{fig:likert} shows the average Likert ratings for the bots for all evaluator groups. 

\subsection{Pairwise Outcomes}
\label{sec:pairwise_outcomes}

\begin{figure}[htb]
\centering
\begin{subfigure}{0.75\columnwidth}
    \includegraphics[width=\columnwidth]{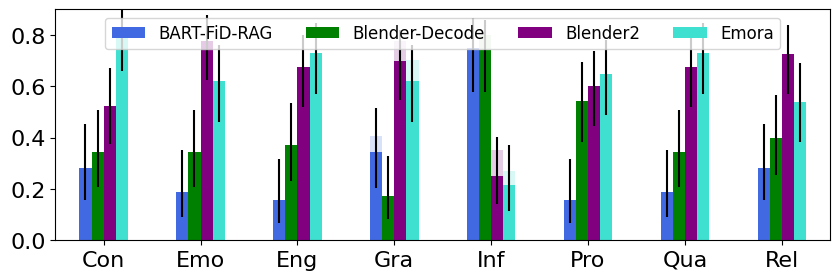}
    \caption{$Dev_x$ evaluations}
    \label{fig:developer_comparative}
\end{subfigure}
\begin{subfigure}{0.75\columnwidth}
    \includegraphics[width=\columnwidth]{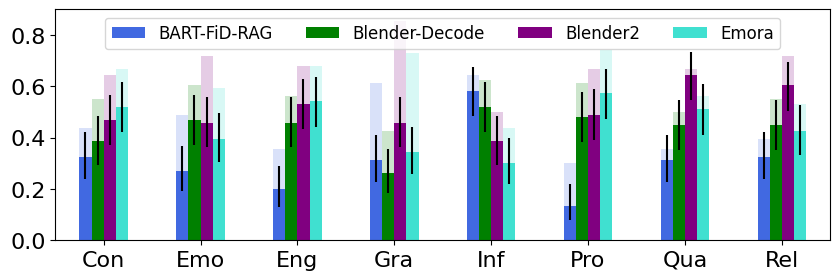}
    \caption{$Stu_i$ evaluations}
    \label{fig:interactive_comparative}
\end{subfigure}
\begin{subfigure}{0.75\columnwidth}
    \includegraphics[width=\columnwidth]{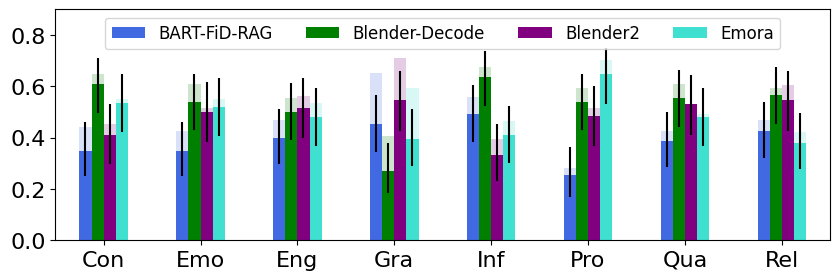}
    \caption{$Stu_x$ evaluations}
    \label{fig:nondeveloper_comparative}
\end{subfigure}
\begin{subfigure}{0.75\columnwidth}
    \includegraphics[width=\columnwidth]{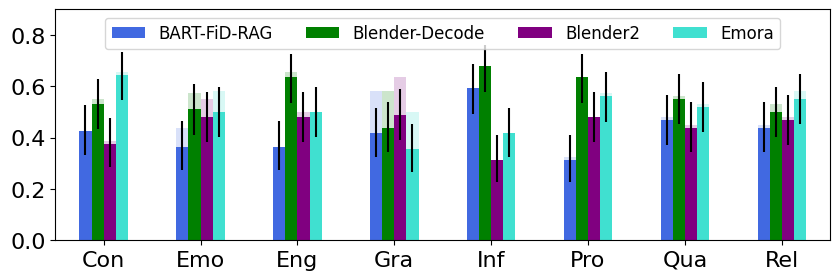}
    \caption{$Sur_x$ evaluations}
    \label{fig:external_comparative}
\end{subfigure}
\caption{Win/Tie proportions for the 4 bots, with 95\% Wilson score confidence intervals for win proportion. Transparent segments denote tie rates.}
\label{fig:comparative}
\end{figure}


Similarly, the holistic overview of Pairwise results is shown in Figure \ref{fig:comparative} for the bots for all evaluator groups. The proportions of wins and ties are measured between each bot and all of the other 3 bots to present an aggregate metric for each bot. 

\end{document}